%% file: main.tex
\documentclass{article}
\usepackage[T1]{fontenc}
\usepackage{spconf}
\usepackage{amsmath,amssymb,amsthm,mathtools}
\usepackage{graphicx,booktabs}
\usepackage{cite}
\usepackage{xcolor}
\usepackage{enumitem}
\usepackage[normalem]{ulem}   

\usepackage[hyphens]{url}
\usepackage[hidelinks]{hyperref}
\graphicspath{{figures/}}

\theoremstyle{plain}
\newtheorem{theorem}{Theorem}
\newtheorem{proposition}[theorem]{Proposition}
\newtheorem{lemma}[theorem]{Lemma}

\theoremstyle{definition}

\theoremstyle{remark}
\newtheorem{remark}{Remark}
\DeclareMathOperator{\msign}{msign}
\DeclareMathOperator{\tr}{tr}
\DeclareMathOperator{\rank}{rank}
\DeclareMathOperator{\diag}{diag}
\newcommand{\R}{\mathbb{R}}
\newcommand{\Fro}[1]{\lVert #1\rVert_{\mathrm F}}
\newcommand{\opnorm}[1]{\lVert #1\rVert_2}

\newcommand{\nuc}[1]{\lVert #1\rVert_{*}}
\newcommand{\ip}[2]{\langle #1,#2\rangle}      
\newif\ifarxiv
\arxivtrue
\newcommand{\extended}{\cite{jawanpuria26a}}

\title{Constant Step Sizes for Muon-Type Spectral Optimizers\\ on Strongly Convex Quadratics}
\name{Pratik Jawanpuria$^{\star}$ \qquad Bhabatosh Garai$^{\star}$ \qquad Bamdev Mishra$^{\dagger}$}
\address{$^{\star}$Centre for Machine Intelligence and Data Science,
         IIT Bombay, India \qquad
         $^{\dagger}$Microsoft, India\\
         {\tt\small \{pratik.jawanpuria, bhabatosh\}@iitb.ac.in} \qquad
         {\tt\small bamdevm@microsoft.com}}

\begin{document}
\maketitle

\begin{abstract}
Muon and related optimizers update a weight matrix with a spectrally transformed gradient. In this work, we study such spectral optimizers with a constant step size on strongly convex matrix quadratics, which isolate the effect of the update from the confounds of nonconvexity, noise, and step size schedules. We analyze an update using two quantities: the gain, which is the factor by which it amplifies the gradient along each singular direction, and the length, which is the norm of the update. These two quantities place the popular spectral updates in three regimes with different behavior: (i) with a bounded gain, the iterates behave like preconditioned gradient descent and converge for step sizes set by the largest curvature; (ii) with an unbounded gain but a vanishing length, they still converge, with the admissible step sizes set by the total curvature instead; and (iii) with a non-vanishing length, they do not converge for almost every initialization. Interestingly, the practical variants of Muon fall in all three regimes, depending only on how the update is normalized and scaled and not on the polynomial that approximates the polar factor. Experiments validate these findings.
\end{abstract}


\section{Introduction}
\label{sec:intro}

Muon~\cite{jordan2024muon} is {a recent} optimizer for the matrix-valued parameters of neural networks, {which has gained popularity}. Instead of treating a weight matrix as a long vector, it replaces the gradient (or momentum) matrix $G$ by an approximation of its polar factor $UV^\top$, where $G=U\Sigma V^\top$ is a singular value decomposition (SVD). The approximation avoids an SVD: the matrix is divided by its Frobenius norm and passed through a few Newton--Schulz (NS) iterations, each of which is a fixed odd matrix polynomial~\cite{jordan2024muon,liu2025moonshot}. Existing literature shows Muon matching AdamW~\cite{loshchilov2018decoupled} on language model pretraining at roughly half of the training compute~\cite{liu2025moonshot}, and it is now used in large-scale training runs.


Several variants of Muon have been explored, and they differ mostly in normalization and scaling (of input/output of NS iterations). For instance, \cite{jordan2024muon,liu2025moonshot} divide the input by its Frobenius norm plus a small constant $\varepsilon$ and multiply the output by a factor that depends only on the shape of the matrix. Steepest descent in the spectral norm~\cite{bernstein2024old} and PolarGrad~\cite{lau2026polargrad} multiply the polar factor by the nuclear norm of the gradient, so that the update vanishes together with the gradient. Polar Express~\cite{amsel2026polar} uses polynomials that approximate the polar factor more accurately at a given depth, and Schatten-$p$ descent~\cite{massena2026schatten} and Freon~\cite{shumaylov2026notspecial} replace the polar factor by a power of the singular values. All of these updates keep the singular vectors of the gradient and change only its singular values. We refer to them as \emph{spectral updates}.


The existing convergence guarantees for Muon are worst-case bounds for general, often nonconvex and stochastic, objectives~\cite{li2025note,kovalev2025understanding,kim2026convergence,li2026finite,shen2025convergence}. With a decreasing step size they give sublinear rates, and with a constant step size they give convergence only up to a neighborhood of the solution. Such bounds cannot tell whether a given update converges at a fixed step size or merely stalls. This distinction matters, since Muon is run with a constant learning rate for most of training, and an exact answer requires a setting that isolates the effect of the update from the confounds of nonconvexity, noise, and step size schedules. The strongly convex matrix quadratic with a constant step size is the simplest such setting, and every effect in it can be traced back to the update. 
Gonon et al.~\cite{gonon2026insights} consider the isotropic quadratic with a constant step size and show that the exact polar update stays on a lattice (very coordinate moves by exactly the step size, so the iterate stays on a grid whose spacing is the step size) and, for almost every initialization, never reaches the minimizer. Shulgin et al.~\cite{shulgin2026quadratic} consider an anisotropic quadratic under exact line search and show that the polar update acts as an iterate-dependent preconditioner, whose amplification of small singular directions is kept bounded by finite NS stages. However, the behavior of spectral updates at a constant step size on anisotropic quadratics, and its dependence on how the updates are normalized and scaled, is not well understood. 

In this work, we propose a single lens through which various spectral updates can be compared, and use it to determine which of them converge at a constant step size on strongly convex matrix quadratics. The lens consists of two quantities. The \emph{gain} is the factor by which the update amplifies the gradient along each singular direction, and the \emph{length} is the overall norm of the update. Gradient descent (GD) has gain one and a length proportional to the gradient, whereas the polar map (vanilla Muon) has gain  $1/\sigma$, which is unbounded as a singular value $\sigma$ tends to zero, and a constant length. Since a bounded gain forces the length to vanish at the minimizer, the two quantities place every spectral update in one of three regimes, and the regime decides its behavior at a constant step size: the length decides whether the iterates converge, and the gain decides how large the step size may be. Our contributions are the following statements about the three regimes.

\begin{itemize}[nosep,leftmargin=*]
\item \textbf{Bounded gain.} Every constant step size below an explicit threshold yields linear convergence (Theorem~\ref{thm:convergence}). Near the minimizer, the iterates behave like GD, so no such update improves on the asymptotic rate of GD (Proposition~\ref{prop:rate}). This regime contains the popular NS iteration of Muon~\cite{jordan2024muon,liu2025moonshot} with its normalization offset.
\item \textbf{Unbounded gain, vanishing length.} The polar map scaled by the nuclear norm of the gradient~\cite{bernstein2024old,lau2026polargrad} converges linearly for any constant step size below a threshold (Theorem~\ref{thm:nuclear}). However, unlike the previous regime, the asymptotic rate depends on initialization.
\item \textbf{Non-vanishing length.} The polar map and the NS polynomial  normalized by the current gradient norm, fail to converge from almost every initialization, at every constant step size (Lemma~\ref{thm:nonconv}). This answers a question of~\cite{gonon2026insights}: the approximate polar factor may not restore convergence.
\end{itemize}
We validate each statement numerically in Section~\ref{sec:numerics}. Proofs and other details are available in our technical report \cite{jawanpuria26a} at \url{https://github.com/Bhabatoshgarai243/spectral-optimizers-quadratics}.

\section{Problem Setting and Spectral Updates}
\label{sec:setting}

We begin with a few notations. For a matrix $G\in\R^{d\times n}$ of rank $r$, let $G=U\Sigma V^\top$ denote its thin SVD, where $U\in\R^{d\times r}$ and $V\in\R^{n\times r}$ have orthonormal columns and $\Sigma=\diag(\sigma_1,\ldots,\sigma_r)$ contains the positive singular values. We write $\Fro{\cdot}$ for the Frobenius norm, $\nuc{\cdot}=\sum_i\sigma_i$ for the nuclear norm, and $\tr(\cdot)$ for the trace. The polar factor of $G$ is $\msign(G):=UV^\top$, with $\msign(0):=0$. 
A \emph{fixed shaping} replaces every singular value $\sigma_i$ of $G$ by $h(\sigma_i)$, where $h:[0,\infty)\to[0,\infty)$ is a scalar function with $h(0)=0$ and $h(\sigma)>0$ for $\sigma>0$, and keeps the singular vectors. We write $h'(0):=\lim_{\sigma\downarrow0}h(\sigma)/\sigma$ for the slope of $h$ at zero, which may be infinite. The choice $h(\sigma)=\sigma$ leaves $G$ unchanged, and $h(\sigma)=1$ for $\sigma>0$ gives the polar factor.

\noindent\textbf{Objective.}
Let $H\in\R^{d\times d}$ be a symmetric positive definite matrix with extreme eigenvalues $0<\lambda_{\min}\le\lambda_{\max}$ and condition number $\kappa=\lambda_{\max}/\lambda_{\min}$. The eigenvalues of $H$ are the curvatures of the objective below, and we call $\lambda_{\max}$ the largest curvature and $\tr(H)$ the total curvature. For a given $X_\star\in\R^{d\times n}$, we consider the matrix quadratic
\begin{equation}
f(X)=\tfrac12\tr\!\left((X-X_\star)^\top H(X-X_\star)\right),\ \ X\in\R^{d\times n}.
\label{eq:objective}
\end{equation}
We write $E_t=X_t-X_\star$ for the error and $G_t=\nabla f(X_t)=HE_t$ for the gradient at iteration $t$, so that $f(X_\star)=0$. 

\noindent\textbf{Spectral updates.}
For a nonzero gradient $G$ with thin SVD $G=U\Sigma V^\top$, a spectral update replaces every singular value $\sigma_i$ by a new value $g_i$ that depends only on the singular values of $G$, and keeps the singular vectors $\Phi(G)=U\diag\big(g_1,\ldots,g_r\big)V^\top$ where $g_i=g_i(\sigma_1,\ldots,\sigma_r)>0$ and $\Phi(0)=0$. The iteration with a constant step size $\alpha>0$ is 
\begin{equation}
X_{t+1}=X_t-\alpha\,\Phi(G_t),
\label{eq:iteration}
\end{equation}
for which $X_\star$ is a fixed point. Table~\ref{tab:instances} lists the instances such as GD, vanilla Muon, Muon via NS iterations~\cite{jordan2024muon}, and PolarGrad~\cite{bernstein2024old,lau2026polargrad} that fall directly under our regimes.  

\begin{table}[t]
\centering
\caption{Instances of the spectral update and their regime. GD has $h(\sigma)=\sigma$ and the polar map has $h\equiv1$. The polynomial $p$ is the composition of the Newton--Schulz (NS) stages.}
\label{tab:instances}
\small
\setlength{\tabcolsep}{4pt}
\begin{tabular}{@{}llll@{}}
\toprule
Update & $g_i$ & Gain $g_i/\sigma_i$ & Regime \\
\midrule
GD & $\sigma_i$ & $1$ & bounded gain \\
Fixed shaping & $h(\sigma_i)$ & $h(\sigma_i)/\sigma_i$ & bounded gain$^\dagger$ \\
NS, $\varepsilon>0$ & $p\big(\tfrac{\sigma_i}{\Fro{G}+\varepsilon}\big)$ & $\le C_p/\varepsilon$ & bounded gain \\
NS, $\varepsilon=0$ & $p\big(\tfrac{\sigma_i}{\Fro{G}}\big)$ & unbounded & non-vanish. length \\
Polar map (Muon) & $1$ & $1/\sigma_i$ & non-vanish. length \\
PolarGrad & $\nuc{G}$ & $\nuc{G}/\sigma_i$ & vanishing length \\
\bottomrule
\multicolumn{4}{@{}l}{\footnotesize $^\dagger$ if $h(\sigma)/\sigma$ is bounded, e.g.\ $h(\sigma)=\tanh\sigma$. $C_p=\max_{0<u\le1}p(u)/u$.}
\end{tabular}
\end{table}


\noindent\textbf{Gain and length.}
Every spectral update is a preconditioned gradient step. Since $U^\top U=I$, we have
\begin{equation}
\Phi(G)=P(G)\,G,\ \
P(G)=U\diag\big(r_1,\ldots,r_r\big)U^\top,
\label{eq:preconditioning}
\end{equation}
where $r_i={g_i}/{\sigma_i}$ is the factor by which the update amplifies the gradient along its $i$-th singular direction, relative to GD. We call $r_i$ the \emph{gain} of the update. The matrix $P(G)$ is symmetric positive semidefinite, its nonzero eigenvalues are the gains, and it is rebuilt from the current gradient at every iteration. The second quantity is the \emph{length} of the update, $\Fro{\Phi(G)}=\big(\sum_ig_i^2\big)^{1/2}$. GD has gain one and length $\Fro{G}$. The polar map has gain $1/\sigma_i$, which grows without bound as a singular value tends to zero, and length $\sqrt r$, which does not depend on the gradient at all.

It should be noted that gain and length quantities are not independent. If every gain is at most $C$, then $\Fro{\Phi(G)}\le C\Fro{G}$, so the length vanishes at the minimizer. Hence, a spectral update falls in one of three regimes (Table~\ref{tab:instances}): a bounded gain, in which case the length vanishes (Section~\ref{sec:bounded}), an unbounded gain with a length that does not vanish (Section~\ref{sec:constant}), and an unbounded gain with a length that vanishes nevertheless (Section~\ref{sec:nuclear}). 

\section{Bounded Gain}
\label{sec:bounded}


We begin with the first regime, in which every gain is bounded and the update is a preconditioned gradient step with a bounded preconditioner. The analysis then follows that of GD, with one addition: a gain bound is only useful if it holds at every gradient the iteration can produce, so we first fix that range. For the strongly convex quadratic \eqref{eq:objective}, as long as the objective does not increase, every gradient satisfies $\Fro{G_t}\le B:=\sqrt{2\lambda_{\max}f(X_0)}$, since $\Fro{G}^2=\tr(E^\top H^2E)\le2\lambda_{\max}f(X)$. The step-size condition below keeps the objective decreasing, so only gradients with $\Fro{G}\le B$ enter the analysis.

\begin{theorem}[Linear convergence under a bounded gain]
\label{thm:convergence}
Let $\Phi$ be a spectral update whose gains satisfy $c\le r_i\le C$ at every $G$ with $0<\Fro{G}\le B$, where $0<c\le C<\infty$. If $0<\alpha C\lambda_{\max}<2$ and $u=\alpha C\lambda_{\max}$, then for every $t\ge0$,
\begin{equation}
f(X_{t+1})\le q\,f(X_t),\ \
q=1-u(2-u)\,\frac{c\,\lambda_{\min}}{C\lambda_{\max}}\in[0,1). 
\label{eq:contraction}
\end{equation}
In particular, $X_t$ converges to $X_\star$ linearly.
\end{theorem}


The condition $\alpha C\lambda_{\max}<2$ is the classical rule $\alpha L<2$ of GD with the smoothness constant ($L$) replaced by $C\lambda_{\max}$. 
While polar map (vanilla Muon) does not fall in this regime, we may consider the fixed shaping $h(\sigma)=\tanh(\sigma/\theta)$. In this case, $C=1/\theta$ and  $\alpha<2\theta/\lambda_{\max}$. We observe that as $\theta\to0$, this map approaches the polar map. 
For the normalized NS iteration with $\varepsilon>0$, the normalized singular values lie in $(0,1)$ and Theorem~\ref{thm:convergence} holds with $C=C_p/\varepsilon$ and $c=c_p/(B+\varepsilon)$ with  $c_p=\min_{0<u\le1}p(u)/u$ for $\alpha C_p\lambda_{\max}<2\varepsilon$. 

Theorem~\ref{thm:convergence} states when a bounded-gain update converges. The next question is whether it converges faster than GD. For GD with step size $\beta$, the error follows the recursion $E_{t+1}=(I-\beta H)E_t$, so $\Fro{E_t}$ decreases asymptotically by the factor $\max_i|1-\beta\lambda_i(H)|$ per step~\cite{polyak1987introduction}. Near the minimizer, every singular value of the gradient is small, and if the gains approach a common limit $a$ there, the update is approximately GD with the effective step size $\alpha a$. The following result shows that the asymptotic rate is then exactly that of GD.
\begin{proposition}[Local rate]
\label{prop:rate}
Under the hypotheses of Theorem~\ref{thm:convergence}, if $r_i\to a>0$ as $\Fro{G}\to0$, uniformly in $i$ and in $G$, then $\limsup_{t\to\infty}\Fro{E_t}^{1/t}\le\rho(\alpha):=\max_i|1-\alpha a\lambda_i(H)|$, and the bound is attained for a suitable initialization.
\end{proposition}
Hence, no bounded-gain update improves on the asymptotic rate of GD, and the best factor any of them can reach is $(\kappa-1)/(\kappa+1)$, attained at $\alpha a=2/(\lambda_{\max}+\lambda_{\min})$. 
For the updates of Table~\ref{tab:instances} in this regime, the limit is $a=h'(0)$ for a fixed shaping and $a=p'(0)/\varepsilon$ for the normalized NS iteration, where $p'(0)=C_p$ for the NS polynomial of Muon~\extended, so the offset $\varepsilon$ acts as a step-size parameter and does not change the achievable rate.


\section{Unbounded Gain with Vanishing Length}
\label{sec:nuclear}

We observe that Theorem~\ref{thm:convergence} needs a bounded gain, and Table~\ref{tab:instances} contains two kinds of updates without one: those with a non-vanishing length, which Section~\ref{sec:constant} treats, and PolarGrad~\cite{lau2026polargrad,bernstein2024old}, which keeps the polar direction but scales it by the nuclear norm of the gradient, $\Phi(G)=\nuc{G}\msign(G)$. The gain $\nuc{G}/\sigma_i$ of this update is unbounded (since a single small singular value receives the full weight $\nuc{G}$), but its length $\nuc{G}\sqrt r$ vanishes with the gradient. The question is whether the vanishing length alone gives convergence at a constant step size, and at which step sizes. 
In this regard, we have the following result. 
\begin{theorem}[Linear convergence under nuclear scaling]
\label{thm:nuclear}
For $\Phi(G)=\nuc{G}\msign(G)$, $0<\alpha\tr(H)<2$, and $u=\alpha\tr(H)$ the iteration \eqref{eq:iteration} satisfies, for every $t\ge0$,
\begin{equation}
\begin{array}{c}
f(X_{t+1})\le\Big(1-u(2-u)\frac{\lambda_{\min}}{\tr(H)}\Big)f(X_t).
\label{eq:nuclear-contraction}
\end{array}
\end{equation}
\end{theorem}
We note that \cite{lau2026polargrad} proves linear convergence of the PolarGrad update on smooth strongly convex objectives with an iterate-dependent step size, whereas Theorem~\ref{thm:nuclear} is a constant step size result. Since the update is homogeneous of degree one in the gradient, its gains $\nuc{G}/\sigma_i$ are the same at every scale, and its behavior at a small error is the same nonlinear function of the direction of the error as at a large one. Hence the gains do not approach a constant near the minimizer, Proposition~\ref{prop:rate} has no analogue here, and the asymptotic rate depends on the initialization \extended.

\begin{figure*}[t]
\centering
\begin{minipage}{0.465\textwidth}
\centering
\includegraphics[width=\linewidth]{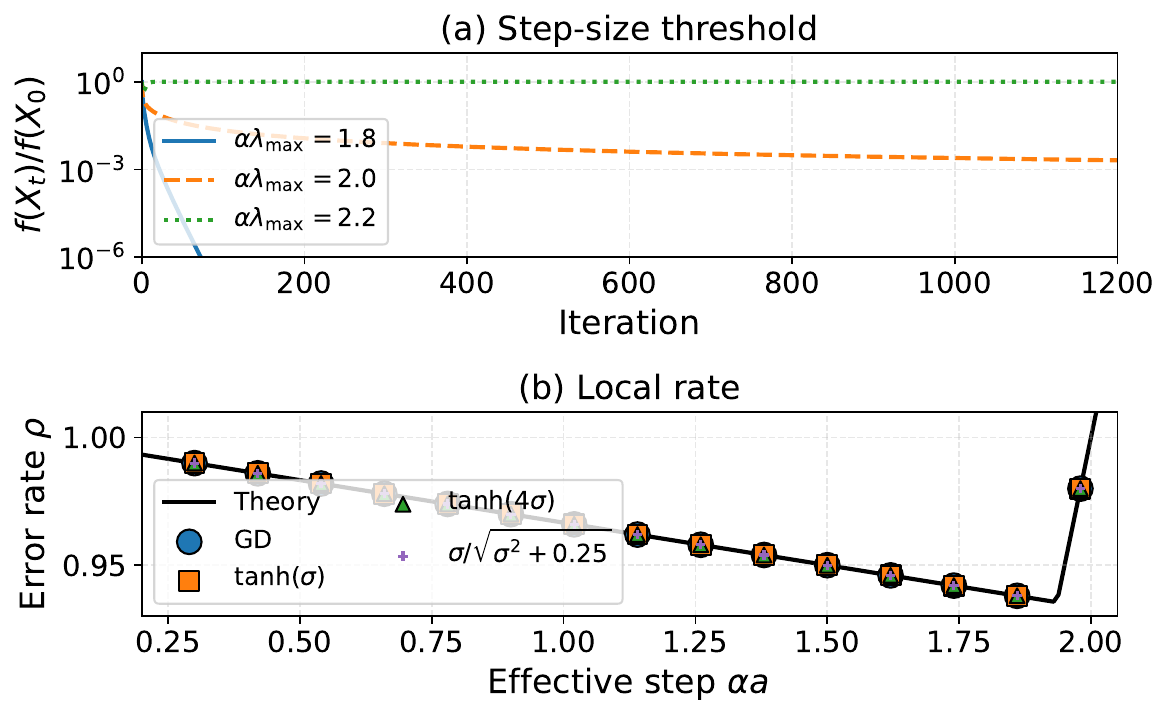}
\end{minipage}\hfill
\begin{minipage}{0.465\textwidth}
\centering
\includegraphics[width=\linewidth]{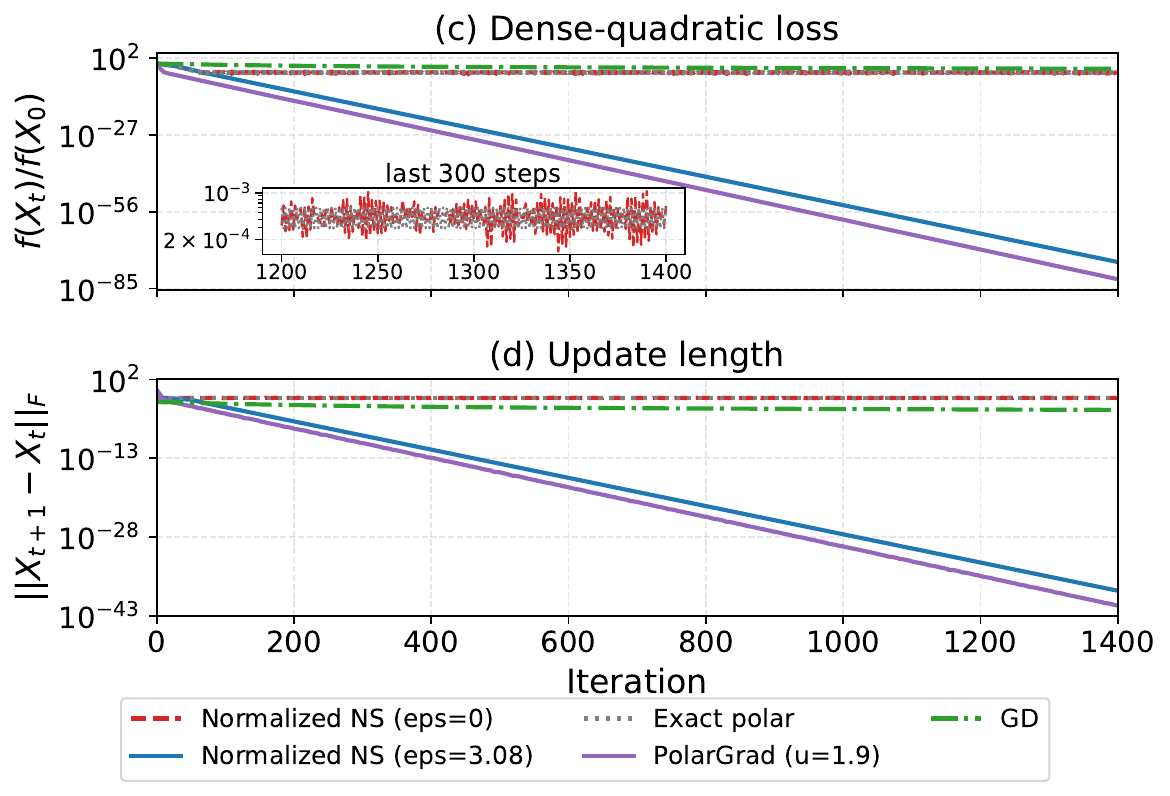}
\end{minipage}
\caption{Left, diagonal problem: (a) P1, $\tanh\sigma$ below, at, and above the threshold. (b) P2, measured error factors of four bounded-gain updates on the curve $\rho(\alpha)$. Right, dense $16\times8$ problem, P3: (c) relative loss and (d) update length of the normalized NS iteration with $\varepsilon=0$ and $\varepsilon=3.08$, the polar map, and GD at the same step size, and PolarGrad at $\alpha\tr(H)=1.9$. }
\label{fig:results}
\end{figure*}

\section{Nonconvergence at Constant Step Size}
\label{sec:constant}

Gonon et al.~\cite{gonon2026insights} show that the exact polar map with a constant step size never enters small neighborhoods of the minimizer on the isotropic quadratic, and ask whether the approximate polar factor that the NS iteration computes changes this. The answer depends on how the input of the NS stages is normalized. Shulgin et al.~\cite{shulgin2026quadratic} divide the input by a constant $s>0$ fixed before the run. This is the fixed shaping $h(\sigma)=p(\sigma/s)$, whose gain is bounded, so it converges linearly by Theorem~\ref{thm:convergence}. We next consider the normalized NS iteration with $\varepsilon=0$, which divides the input by the current gradient norm, as a candidate with approximate polar factor.

\begin{lemma}[Nonconvergence]
\label{thm:nonconv}
Let $\Phi$ be the polar map or the normalized NS iteration with $\varepsilon=0$. Then for every $H\succ0$, $\alpha>0$, and $X_0$, the iteration \eqref{eq:iteration} satisfies $\max\{\Fro{E_t},\Fro{E_{t+1}}\}\ge\alpha\gamma/2$ whenever $G_t\ne0$, where $\gamma=1$ for the polar map and $\gamma=c_p$ for the NS iteration, and the iterates have no limit for almost every $X_0$.
\end{lemma}

The lemma extends the result of~\cite{gonon2026insights} from the isotropic Hessian to every Hessian, and from the exact polar map to the NS polynomial of every degree. The approximate polar factorization therefore does not restore convergence by itself. Normalization also influences convergence: the same polynomial converges linearly when its input is divided by a fixed constant or by the gradient norm plus an offset ($\varepsilon>0$ in Table~\ref{tab:instances}), and fails at every constant step size when it is divided by the gradient norm alone. This is consistent with the observation of~\cite{gonon2026insights} that Gaussian noise on the sign restores the reachability of small neighborhoods: noise breaks the grid, but the perturbed update still has a non-vanishing length and does not converge, whereas the offset or the fixed scale bounds the gain and does.

\section{Numerical Experiments}
\label{sec:numerics}

We test the three regimes and the local rate on small problems, checking in each case that the behavior at the stated condition, and just beyond it, is the predicted one, since the thresholds of Theorems~\ref{thm:convergence} and~\ref{thm:nuclear} are only sufficient and the rate of Proposition~\ref{prop:rate} is only an upper bound. All runs minimize \eqref{eq:objective} with $X_\star=0$ and $\kappa=30$ at a constant step size, on a diagonal problem with $d=n=16$ (P1, P2) or a dense problem with $d=16$, $n=8$ (P3, P4). Seeds and recorded values are given in \extended.

\noindent\textbf{P1: bounded gain, step-size threshold.}
For $h(\sigma)=\tanh\sigma$ the gain bound is $C=1$, so Theorem~\ref{thm:convergence} guarantees linear convergence exactly for $\alpha\lambda_{\max}<2$. Fig.~\ref{fig:results}(a) shows the runs at $\alpha\lambda_{\max}\in\{1.8,2.0,2.2\}$: linear convergence below the threshold, sublinear convergence at equality, with a loss of order $1/t$ \extended, and a persistent cycle above it. 

\noindent\textbf{P2: local rate.}
We run GD and three fixed shapings with different slopes at zero, $\tanh\sigma$, $\tanh4\sigma$, and $\sigma/\sqrt{\sigma^2+0.25}$, at 15 values of the effective step size $\alpha a$ and measure the asymptotic factor of $\Fro{E_t}$ per step. Although the four updates act very differently on large singular values, their measured factors fall on the single curve $\rho(\alpha)$ of Proposition~\ref{prop:rate} in Fig.~\ref{fig:results}(b), with a maximum deviation below $3\times10^{-8}$. The minimum of the curve is the optimal rate of GD, $(\kappa-1)/(\kappa+1)\approx0.935$, on which none of the updates improves.

\noindent\textbf{P3: the three regimes on one problem.}
We run the normalized NS iteration with $\varepsilon=0$ and with $\varepsilon=3.08$, the polar map, and GD from the same $E_0$ at the same step size, and PolarGrad at $\alpha\tr(H)=1.9$. The offset $\varepsilon$ is chosen so that the shared step size satisfies $\alpha C_p\lambda_{\max}/\varepsilon=1.8$, inside the range of Theorem~\ref{thm:convergence}, and it is large enough as that range shrinks with $\varepsilon$. Fig.~\ref{fig:results}(d) shows the mechanism: the update length shrinks for the updates in the convergent regimes and stays bounded below for the polar map and the NS iteration with $\varepsilon=0$. Fig.~\ref{fig:results}(c) shows the consequence: the latter two oscillate at a relative loss between $10^{-4}$ and $10^{-3}$, whereas the NS iteration with $\varepsilon=3.08$ and PolarGrad converge linearly to optimal solution. The two NS runs share the polynomial and differ only in $\varepsilon$, so the normalization alone separates the regimes. PolarGrad also converges at $\alpha\tr(H)=1.0$ and no longer converges at $\alpha\tr(H)=2.4$, in line with the threshold of Theorem~\ref{thm:nuclear}.


\noindent\textbf{P4: the offset as a step-size parameter.}
We run the normalized NS iteration near the minimizer at nine pairs $(\varepsilon,\alpha)$ with $\varepsilon$ over a tenfold range and effective step size $\alpha p'(0)\lambda_{\max}/\varepsilon\in\{1.0,1.9,2.05\}$. Runs with the same effective step size have the same measured factor to machine precision, matching $\rho(\alpha)$ of Proposition~\ref{prop:rate} to $2.4\times10^{-5}$, and only the run above $2$ grows. Hence, $\varepsilon$ affects the local rate only through the effective step size.


\section{Conclusion}
\label{sec:conclusions}

We studied spectral updates with a constant step size on strongly convex matrix quadratics through two quantities, the gain and the length of the update. They place the popular Muon-style spectral updates in three regimes: a bounded gain gives linear convergence at the rate of GD, a vanishing length with an unbounded gain gives linear convergence under a step size set by the total curvature, and a non-vanishing length gives no convergence at all. Extending the analysis to momentum and stochastic gradients is left for future work.
\bibliographystyle{IEEEbib}
\bibliography{ref-2}
\clearpage

 \input{appendix}

\end{document}

%% file: appendix.tex
\onecolumn
\appendix

\noindent\textbf{Organization of the appendix.}
Appendix~\ref{app:related} relates our results to prior work. Appendices~\ref{app:convergence}, \ref{app:nuclear}, and~\ref{app:nonconvergence} contain the proofs for the three regimes: the bounded gain (Theorem~\ref{thm:convergence} and Proposition~\ref{prop:rate}), the nuclear scaling (Theorem~\ref{thm:nuclear}), and the non-vanishing length (Lemma~\ref{thm:nonconv}). Appendix~\ref{app:boundary} treats the boundary of the step-size range of Theorem~\ref{thm:convergence} and shows that a vanishing length alone does not give convergence. Appendix~\ref{app:muon} computes the constants of the Newton--Schulz polynomial of Muon, and Appendix~\ref{app:experiments} lists the details of the experiments.

\section{Relation to prior results}
\label{app:related}

\noindent\textbf{Constant step sizes on quadratics.}
Gonon et al.~\cite{gonon2026insights} show on $\frac12\Fro{W}^2$ that exact polar iterates with a constant step size stay on the lattice $s_0+\alpha\mathbb Z$ coordinate by coordinate, end in a two-cycle with a loss of at most $\alpha^2/2$ per coordinate, and require $\Theta(1/\sqrt\epsilon)$ steps to reach a loss of $\epsilon$ with a tuned constant step. They ask whether the approximate polar factor computed by the Newton--Schulz iteration changes the reachability of small neighborhoods, prove that additive Gaussian noise on the scalar sign map restores it, and study the actual Newton--Schulz iteration in experiments, where the approximation can change which method is faster. Lemma~\ref{thm:nonconv} answers the deterministic question: it covers every positive definite $H$, concludes the absence of a limit, and shows that the finite polynomial with the per-step normalization, which divides the input by the current gradient norm ($\varepsilon=0$ in Table~\ref{tab:instances}), fails at every degree.

\noindent\textbf{Exact line search on the same quadratic.}
Shulgin et al.~\cite{shulgin2026quadratic} study the objective \eqref{eq:objective} under exact line search. They write the polar step as $A(X)X$ with $A(X)=(HXX^\top H)^{\dagger/2}H$ (for $X_\star=0$, with $\dagger/2$ the square root of the pseudoinverse), show that the line-search step size is homogeneous of degree one in $X$ and hence vanishes at the minimizer, and analyze damped polar maps $\sigma/\sqrt{\sigma^2+\epsilon}$, rank truncation, and finite Newton--Schulz stages. For Newton--Schulz stages with the input divided by a fixed constant, they prove that the multiplier on every singular direction is bounded by $C_p/s$ in our notation, where $s$ is that constant and $p$ is the composed polynomial, and that $p(u)=p'(0)u+O(u^3)$, where $p'(0)$ is the product of the linear coefficients of the stages. This multiplier is the gain of Section~\ref{sec:setting} evaluated at a fixed input scale, and Lemma~\ref{lem:instances}(c) records that this update is a fixed shaping in our sense, so that Theorem~\ref{thm:convergence} applies to it. What the present paper adds is the analysis at a constant step size: the step-size threshold with its constant (Theorem~\ref{thm:convergence}), the range $B$ of the gradients that the iteration reaches (Section~\ref{sec:bounded}), the failure of the per-step normalization (Lemma~\ref{thm:nonconv}), the role of $\varepsilon$ and the local rate (Proposition~\ref{prop:rate}), and the trace threshold for the nuclear scaling (Theorem~\ref{thm:nuclear}).

\noindent\textbf{Null-gradient consistency and nuclear scaling.}
PolarGrad~\cite{lau2026polargrad} calls an optimizer null-gradient consistent when its update size tends to zero as the gradient tends to zero, observes that Muon lacks this property, and restores it by scaling the polar factor by the nuclear norm of the gradient. In our terms, null-gradient consistency is a vanishing length, and a bounded gain implies it since $\Fro{\Phi(G)}\le C\Fro{G}$. PolarGrad also proves linear convergence of the nuclear-scaled polar update on smooth strongly convex objectives with the iterate-dependent step size $\gamma_k=1/(Lr_k)$, $r_k=\rank(G_k)$, at the rates $1-1/(r_k\kappa)$ and $1-1/(\kappa_{G_k}^2\kappa)$, where $\kappa_{G_k}$ is the condition number of $G_k$. It also shows that matrix sign descent with a constant step size satisfies a recursion that plateaus at a floor, and it remarks that the constant step size $1/(Lr_{\max})$, where $r_{\max}$ is the largest rank along the trajectory, gives the uniform rate $O(\exp(-k/(r_{\max}\kappa)))$. On the quadratic, Theorem~\ref{thm:nuclear} gives the threshold $\alpha\tr(H)<2$ for a constant step size and the rate $1-u(2-u)\lambda_{\min}/\tr(H)$ in closed form from the exact identity \eqref{eq:nuclear-identity}, and Proposition~\ref{prop:invariant} shows that its refinement $\alpha\tr_n(H)<2$ of Appendix~\ref{app:nuclear} is exact, where $\tr_n(H)\le\tr(H)$ is the sum of the $n$ largest eigenvalues of $H$. Here $L=\lambda_{\max}$, a generic trajectory has $r_{\max}=\min(d,n)$, and $\tr_n(H)\le\min(d,n)\lambda_{\max}$, so the range $\alpha\tr_n(H)<2$ contains the constant step size $1/(Lr_{\max})$ of PolarGrad with a margin of at least a factor of two. Steepest descent in the spectral norm~\cite{bernstein2024old} returns the same update, and the spectral-norm oracle of~\cite{pethick2025training} returns $-\rho UV^\top$ with a magnitude set by the radius $\rho$ of the ball, which is an update with a non-vanishing length.

\noindent\textbf{One-step maximal learning rate.}
Nguyen et al.~\cite{nguyen2026spectral} write the exact polar step as preconditioned GD and derive from a second-order Taylor bound the one-step condition $\eta<(2/\lambda_{\max})\sum_i\sigma_i/m$ for the loss to decrease, where $m$ is the number of singular values of the gradient, so that the threshold is $2/\lambda_{\max}$ times their average. On the quadratic \eqref{eq:objective}, Lemma~\ref{lem:identity} gives the exact condition $\alpha\tr(U_t^\top HU_t)<2\nuc{G_t}$ for the polar step, and the bound $\tr(U_t^\top HU_t)\le r\lambda_{\max}$, where $r$ is the rank of $G_t$, recovers theirs with $m=r$. This threshold depends on the singular values of the current gradient, and it is a per-iteration statement for the polar map. Theorem~\ref{thm:convergence} fixes the range of singular values in advance and states a single condition for the whole trajectory and for every bounded-gain update.

\noindent\textbf{Newton--Schulz in convergence analyses.}
If the polynomial $p$ is the composition of $k$ copies of a stage polynomial $q$, then $p'(0)=q'(0)^k$ by the chain rule, since $q(0)=0$. A stage that pushes small singular values towards one has $q'(0)>1$ (for the quintic of~\cite{jordan2024muon}, $q'(0)=3.4445$), so every added stage multiplies the gain near zero and shrinks the guaranteed step size of Theorem~\ref{thm:convergence} for the normalized NS iteration with $\varepsilon>0$. Kim and Oh~\cite{kim2026convergence} prove nonconvex convergence of Muon with $k$ Newton--Schulz stages at a step size built from the horizon, measure stationarity in the nuclear norm, and show that the gap to the exact polar variant closes doubly exponentially in $k$. Their input scaling $\max\{1,\Fro{M}\}$ of the momentum $M$ equals one for $\Fro{M}\le1$, so near a minimizer it is a fixed scale. Li and Tsuchiya~\cite{li2026finite} view Muon with momentum and finite Newton--Schulz as an online learner with a smoothed spectral potential and prove stationarity guarantees for nonsmooth nonconvex objectives. For the classical Newton--Schulz coefficients, the Lipschitz constant of the depth-$k$ map is $(15/8)^k$ while its distance to the polar map decays like $(15/8)^{-k}$. Their input is divided by a constant bound on the gradient norm fixed before the run, and they state that the data-dependent normalization of practical Muon can violate their scale condition. Lemma~\ref{thm:nonconv} shows on the quadratic that the per-step normalization does fail at a constant step size. Polar Express~\cite{amsel2026polar} chooses the stage polynomials to minimize the worst-case distance to the polar factor on an interval $[l,u]$ and rescales the input by its Frobenius norm plus a small offset ($10^{-2}$ in their algorithm). Uniform accuracy on $[l,u]$ and a bounded gain are different requirements, and with the per-step normalization the Polar Express update is in the regime of a non-vanishing length. As for Muon (Appendix~\ref{app:muon}), the small offset bounds the gain only formally.

\noindent\textbf{Sign methods and other spectral families.}
signSGD keeps only the sign of every coordinate~\cite{bernstein2018signsgd}, which is the coordinatewise analogue of setting the singular values to one. It fails at any fixed step size on a one-dimensional convex stochastic example with bimodal gradient noise, because the sign discards the magnitude of the gradient, and error feedback repairs it by reinjecting the discarded magnitude~\cite{karimireddy2019error}. Lemma~\ref{thm:nonconv} is a deterministic matrix counterpart of that failure, in which the polar map discards the magnitudes of all singular values. Schatten-$p$ steepest descent returns a power of the singular values divided by a norm of the gradient~\cite{massena2026schatten}, and Freon~\cite{shumaylov2026notspecial} extends the power into the quasi-norm regime and scales every update by the inner product between the gradient and the shaped direction, which vanishes with the gradient. Without that scaling, a power $\sigma^\gamma$ with $\gamma<1$ has an unbounded gain and a vanishing length, and Proposition~\ref{prop:power} (Appendix~\ref{app:boundary}) shows that it still does not converge, already in the diagonal model, so a vanishing length alone is not sufficient. Shampoo~\cite{gupta2018shampoo} preconditions with accumulated factors that carry memory of past gradients, so its update is not a function of the current gradient alone and lies outside the class of spectral updates. Without accumulation, it reduces to the polar map~\cite{bernstein2024old}.

\section{Proofs for the bounded-gain regime}
\label{app:convergence}

This appendix proves Theorem~\ref{thm:convergence} and Proposition~\ref{prop:rate}. Lemma~\ref{lem:identity} expresses the change of the objective in one step exactly, Lemma~\ref{lem:gain} bounds its two terms through the gains, and Lemma~\ref{lem:instances} verifies the gain bounds for the instances of Table~\ref{tab:instances}. We write $\ip{A}{B}=\tr(A^\top B)$. Three facts about the quadratic \eqref{eq:objective} recur in the proofs. With $E=X-X_\star$ and $G=HE$, one has $2f(X)=\ip{E}{HE}$ and $\Fro{G}^2=\ip{E}{H^2E}$, and the inequalities $H\succeq\lambda_{\min}I$ and $\lambda_{\min}H\preceq H^2\preceq\lambda_{\max}H$ give
\begin{equation}
\Fro{E}^2\le\frac{2f(X)}{\lambda_{\min}},\qquad
2\lambda_{\min}f(X)\le\Fro{G}^2\le2\lambda_{\max}f(X).
\label{eq:three}
\end{equation}
The upper bound on $\Fro{G}^2$ gives the range $B$ of Section~\ref{sec:bounded}, the lower bound on $\Fro{G}^2$ converts a decrease proportional to $\Fro{G}^2$ into a contraction of the objective, and the bound on $\Fro{E}^2$ turns this contraction into one of the error.

\begin{lemma}[One-step identity]
\label{lem:identity}
For any update $X_{t+1}=X_t-\alpha\Phi$ with $\Phi\in\R^{d\times n}$,
\begin{equation}
f(X_{t+1})=f(X_t)-\alpha\ip{G_t}{\Phi}+\tfrac{\alpha^2}{2}\ip{\Phi}{H\Phi}.
\label{eq:expansion}
\end{equation}
\end{lemma}

\begin{proof}
Expand $\tfrac12\ip{E_t-\alpha\Phi}{H(E_t-\alpha\Phi)}$ and use $HE_t=G_t$.
\end{proof}

\begin{lemma}[Gain bounds]
\label{lem:gain}
Let $\Phi(G)=P(G)G$ as in \eqref{eq:preconditioning}, and suppose that at some $G$ every gain satisfies $c\le r_i\le C$. Then
\[
\ip{G}{\Phi(G)}\ge c\,\Fro{G}^2\qquad\text{and}\qquad\ip{\Phi(G)}{H\Phi(G)}\le C\lambda_{\max}\,\ip{G}{\Phi(G)}.
\]
\end{lemma}

\begin{proof}
Since $U^\top G=\Sigma V^\top$, we have $\ip{G}{P(G)G}=\sum_ir_i\sigma_i^2\ge c\sum_i\sigma_i^2=c\Fro{G}^2$. Next, $H\preceq\lambda_{\max}I$ gives $\ip{PG}{HPG}\le\lambda_{\max}\Fro{PG}^2=\lambda_{\max}\ip{G}{P^2G}$, and $P^2=U\diag(r_i^2)U^\top\preceq C\,U\diag(r_i)U^\top=CP$ because $0\le r_i\le C$ implies $r_i^2\le Cr_i$. Hence $\ip{G}{P^2G}\le C\ip{G}{PG}=C\ip{G}{\Phi(G)}$, which gives the second inequality.
\end{proof}

\begin{proof}[Proof of Theorem~\ref{thm:convergence}]
We show by induction that $f(X_t)\le f(X_0)$ and that \eqref{eq:contraction} holds for every $t$. Suppose $f(X_t)\le f(X_0)$. If $G_t=0$, then $X_{t+1}=X_t=X_\star$ and there is nothing to prove. Otherwise, $\Fro{G_t}^2\le2\lambda_{\max}f(X_t)\le B^2$ by \eqref{eq:three}, so the gain bounds hold at $G_t$ and Lemmas~\ref{lem:identity} and~\ref{lem:gain} give, with $u=\alpha C\lambda_{\max}\in(0,2)$,
\[
f(X_{t+1})\le f(X_t)-\alpha\Big(1-\frac u2\Big)\ip{G_t}{\Phi(G_t)}
\le f(X_t)-\alpha\Big(1-\frac u2\Big)c\Fro{G_t}^2
\le\Big(1-2\alpha c\lambda_{\min}\Big(1-\frac u2\Big)\Big)f(X_t),
\]
where the last step uses $\Fro{G_t}^2\ge2\lambda_{\min}f(X_t)$. In particular $f(X_{t+1})\le f(X_t)$, which closes the induction. Substituting $\alpha=u/(C\lambda_{\max})$ gives $2\alpha c\lambda_{\min}(1-u/2)=u(2-u)c\lambda_{\min}/(C\lambda_{\max})$, which is the factor in \eqref{eq:contraction}. It lies in $(0,1]$ because $u(2-u)\le1$, $c\le C$, and $\lambda_{\min}\le\lambda_{\max}$, hence $q\in[0,1)$. Iterating, $f(X_t)\le q^tf(X_0)$, and \eqref{eq:three} gives $\Fro{E_t}^2\le2q^tf(X_0)/\lambda_{\min}$.
\end{proof}

\begin{lemma}[Gain bounds for the instances]
\label{lem:instances}
Let $0<\Fro{G}\le B$. Every gain of the following updates lies in $[c,C]$:
\begin{enumerate}[label=(\alph*),nosep,leftmargin=*]
\item a fixed shaping $h$ that is continuous and positive on $(0,B]$ and has $h'(0)\in(0,\infty)$, with $c=\inf_{0<\sigma\le B}h(\sigma)/\sigma>0$ and $C=\sup_{0<\sigma\le B}h(\sigma)/\sigma<\infty$ (in particular, $C=1/\theta$ for $h(\sigma)=\tanh(\sigma/\theta)$);
\item the normalized NS iteration with $\varepsilon>0$, where $p$ is an odd polynomial that is positive on $(0,1]$ and has $p'(0)>0$, with $c=c_p/(B+\varepsilon)$ and $C=C_p/\varepsilon$;
\item the NS iteration with the same $p$ and the input divided by a constant $s\ge B$ fixed before the run, with $c=c_p/s$ and $C=C_p/s$.
\end{enumerate}
\end{lemma}

\begin{proof}[Proof of Lemma~\ref{lem:instances}]
\emph{(a)} For a fixed shaping $g_i=h(\sigma_i)$ with $h$ continuous and positive on $(0,B]$ and $\lim_{\sigma\downarrow0}h(\sigma)/\sigma=a\in(0,\infty)$, the function $\sigma\mapsto h(\sigma)/\sigma$ extends continuously to the compact interval $[0,B]$ and is positive on it, so $c=\inf_{0<\sigma\le B}h(\sigma)/\sigma>0$ and $C=\sup_{0<\sigma\le B}h(\sigma)/\sigma<\infty$, and every gain at a gradient with $\Fro{G}\le B$ lies in $[c,C]$ since $\sigma_i\le\Fro{G}$. For the soft polar map $h(\sigma)=\tanh(\sigma/\theta)$, the gain $\tanh(\sigma/\theta)/\sigma$ is decreasing in $\sigma$ with limit $1/\theta$ at zero, so $C=1/\theta$.

\emph{(b)} For the normalized NS iteration with $\varepsilon>0$, put $u_i=\sigma_i/(\Fro{G}+\varepsilon)\in(0,1)$. The gain is $r_i=p(u_i)/\sigma_i=\big(p(u_i)/u_i\big)/(\Fro{G}+\varepsilon)$. Since $p$ is an odd polynomial with $p'(0)>0$ and $p>0$ on $(0,1]$, the function $p(u)/u$ extends continuously to $[0,1]$ with value $p'(0)$ at $0$ and is positive there, so $c_p=\inf_{0<u\le1}p(u)/u>0$ and $C_p=\sup_{0<u\le1}p(u)/u<\infty$. Table~\ref{tab:instances} writes them as a minimum and a maximum, but for the polynomial of Muon the supremum $C_p=p'(0)$ is approached only as $u\downarrow0$ (Appendix~\ref{app:muon}). For $\Fro{G}\le B$ we have $\varepsilon\le\Fro{G}+\varepsilon\le B+\varepsilon$, hence $c_p/(B+\varepsilon)\le r_i\le C_p/\varepsilon$.

\emph{(c)} For the NS iteration with the input divided by a constant $s>0$ fixed before the run, as in~\cite{shulgin2026quadratic}, the update is the fixed shaping $h(\sigma)=p(\sigma/s)$. By Appendix~\ref{app:muon}, $p>0$ on $(0,\infty)$, and $h'(0)=p'(0)/s\in(0,\infty)$, so part~(a) applies with $C=s^{-1}\sup_{0<u\le B/s}p(u)/u$ and $c=s^{-1}\inf_{0<u\le B/s}p(u)/u$. If $s\ge B$, so that the input lies in the unit ball, then $C\le C_p/s$ (with equality for the polynomial of Muon, Appendix~\ref{app:muon}) and $c\ge c_p/s$, which uses only that $p>0$ on $(0,1]$, and Theorem~\ref{thm:convergence} gives linear convergence for $\alpha C_p\lambda_{\max}<2s$. The fixed scale $s$ thus plays the role of the offset $\varepsilon$.
\end{proof}

\begin{proof}[Proof of Proposition~\ref{prop:rate}]
\emph{Upper bound.} Fix $\eta>0$. By the uniform convergence of the gains there is $\delta>0$ with $|r_i(G)-a|\le\eta$ for all $i$ whenever $0<\Fro{G}\le\delta$. For such $G$, by unitary invariance of the Frobenius norm,
\[
\Fro{\Phi(G)-aG}^2=\sum_i(r_i-a)^2\sigma_i^2\le\eta^2\Fro{G}^2.
\]
Hence $E_{t+1}=(I-\alpha aH)E_t+R_t$ with $\Fro{R_t}\le\alpha\eta\Fro{G_t}\le\alpha\eta\lambda_{\max}\Fro{E_t}$ whenever $\Fro{G_t}\le\delta$. The operator norm of $E\mapsto(I-\alpha aH)E$ on $(\R^{d\times n},\Fro\cdot)$ is the spectral radius of the symmetric matrix $I-\alpha aH$, which is $\rho(\alpha)$. By Theorem~\ref{thm:convergence}, $E_t\to0$, so $\Fro{G_t}\le\delta$ for all large $t$, and then $\Fro{E_{t+1}}\le(\rho(\alpha)+\alpha\eta\lambda_{\max})\Fro{E_t}$. Hence $\limsup_t\Fro{E_t}^{1/t}\le\rho(\alpha)+\alpha\eta\lambda_{\max}$, and since $\eta$ is arbitrary, $\limsup_t\Fro{E_t}^{1/t}\le\rho(\alpha)$.

\emph{Attainment.} Let $\lambda$ be an eigenvalue of $H$ with $|1-\alpha a\lambda|=\rho(\alpha)$, let $Hw=\lambda w$ with $\|w\|=1$, and let $v\in\R^n$ be a unit vector. For $E=xwv^\top$ with $x\ne0$, the gradient $G=\lambda xwv^\top$ has the single singular value $\lambda|x|$ with singular vectors $\mathrm{sign}(x)w$ and $v$, so $\Phi(G)=g_1\,\mathrm{sign}(x)wv^\top$ and the error stays of the form $x_twv^\top$ with
\[
x_{t+1}=x_t\big(1-\alpha\lambda\,r_1(G_t)\big)=:x_t\,\mu(x_t).
\]
As $x\to0$, $\mu(x)\to1-\alpha a\lambda$, whose modulus is $\rho(\alpha)<1$, since $a\le C$ gives $\alpha a\lambda<2$. Assume $\rho(\alpha)>0$ and choose $|x_0|$ so small that $0<|\mu(x)|<1$ for all $0<|x|\le|x_0|$, which is possible because $\mu$ has a nonzero limit at $0$. Then $|x_t|$ decreases and never vanishes, it tends to $0$ by Theorem~\ref{thm:convergence}, and $|x_{t+1}|/|x_t|=|\mu(x_t)|\to\rho(\alpha)$. Finally $\Fro{E_t}=|x_t|$. If $\rho(\alpha)=0$ the bound is attained trivially.

\emph{The instances.} It remains to check the uniform convergence of the gains for the instances of Section~\ref{sec:bounded}. For a fixed shaping, $r_i=h(\sigma_i)/\sigma_i\to h'(0)=a$ uniformly in $i$ and in $G$ as $\Fro{G}\to0$, since $\sigma_i\le\Fro{G}$. For the normalized NS iteration with $\varepsilon>0$, $r_i=\big(p(u_i)/u_i\big)/(\Fro{G}+\varepsilon)$ with $u_i\le\Fro{G}/\varepsilon\to0$ uniformly in $i$ and in $G$, so $p(u_i)/u_i\to p'(0)$ and $r_i\to p'(0)/\varepsilon=a$. A bounded gain alone does not imply such a limit: the fixed shaping $h(\sigma)=\sigma(2+\sin\ln\sigma)$ has gain in $[1,3]$ and converges by Theorem~\ref{thm:convergence}, but its gain has no limit at zero.
\end{proof}

\begin{remark}[Optimal step size]
For an effective step size $\beta=\alpha a$, $\rho(\alpha)=\max\{|1-\beta\lambda_{\min}|,|1-\beta\lambda_{\max}|\}$ is minimized at $\beta_\star=2/(\lambda_{\min}+\lambda_{\max})$ with value $(\kappa-1)/(\kappa+1)$. Substituting $\alpha=\beta_\star/a$ into the condition $\alpha C\lambda_{\max}<2$ shows that the optimal rate is reachable within the guaranteed range if and only if $C/a<1+1/\kappa$, which holds for concave shapings such as $\tanh(\sigma/\theta)$, where $C=a=1/\theta$, and for the normalized NS iteration of Muon, where $C=a=C_p/\varepsilon$ (Appendix~\ref{app:muon}).
\end{remark}

\section{Proofs for the nuclear scaling}
\label{app:nuclear}

Let $\tr_n(H)$ denote the sum of the $n$ largest eigenvalues of $H$, so that $\tr_n(H)=\tr(H)$ when $n\ge d$ and $\tr_n(H)<\tr(H)$ otherwise. The proof below gives Theorem~\ref{thm:nuclear} with $\tr(H)$ replaced by $\tr_n(H)$, which is a stronger statement, and Proposition~\ref{prop:invariant} shows that the threshold $\alpha\tr_n(H)<2$ cannot be relaxed.

\begin{proof}[Proof of Theorem~\ref{thm:nuclear}]
If $G_t=0$, then $X_{t+1}=X_t=X_\star$ and there is nothing to prove, so let $G_t\ne0$. Write $G=U\Sigma V^\top$ and $\Phi=\nuc{G}UV^\top$. Using $V^\top V=I_r$ and $U^\top G=\Sigma V^\top$,
\[
\ip{G}{\Phi}=\nuc{G}\tr(V\Sigma U^\top UV^\top)=\nuc{G}\tr(\Sigma)=\nuc{G}^2,\qquad
\ip{\Phi}{H\Phi}=\nuc{G}^2\tr(VU^\top HUV^\top)=\nuc{G}^2\tr(U^\top HU).
\]
Substituting into Lemma~\ref{lem:identity} gives the exact identity
\begin{equation}
f(X_{t+1})=f(X_t)-\alpha\nuc{G_t}^2+\tfrac{\alpha^2}{2}\nuc{G_t}^2\tr(U_t^\top HU_t),
\label{eq:nuclear-identity}
\end{equation}
where $U_t$ holds the left singular vectors of $G_t$. The columns of $U_t$ are $r\le\min(d,n)$ orthonormal vectors, so by Ky Fan's maximum principle $\tr(U_t^\top HU_t)$ is at most the sum of the $r$ largest eigenvalues of $H$, which is at most $\tr_n(H)\le\tr(H)$. Assume that $0<\alpha\tr_n(H)<2$. With $u=\alpha\tr_n(H)\in(0,2)$,
\[
f(X_{t+1})\le f(X_t)-\alpha\Big(1-\frac u2\Big)\nuc{G_t}^2\le f(X_t)-\alpha\Big(1-\frac u2\Big)\Fro{G_t}^2\le\Big(1-2\alpha\lambda_{\min}\Big(1-\frac u2\Big)\Big)f(X_t),
\]
using $\nuc{G}\ge\Fro{G}$ and \eqref{eq:three}. Substituting $\alpha=u/\tr_n(H)$ gives \eqref{eq:nuclear-contraction} with $\tr_n(H)$ in place of $\tr(H)$, and the factor lies in $[0,1)$ since $u(2-u)\le1$ and $\lambda_{\min}\le\tr_n(H)$. The statement with $\tr(H)$ follows because $\tr_n(H)\le\tr(H)$, so $\alpha\tr(H)<2$ implies $\alpha\tr_n(H)<2$, and the factor $1-u(2-u)\lambda_{\min}/\tr_n(H)$ at $u=\alpha\tr_n(H)$ is at most $1-u'(2-u')\lambda_{\min}/\tr(H)$ at $u'=\alpha\tr(H)$, since $x\mapsto\alpha x(2-\alpha x)/x=\alpha(2-\alpha x)$ is decreasing in $x$.
\end{proof}

\begin{proposition}[Invariant directions]
\label{prop:invariant}
Let $S$ be a set of at most $n$ eigenvalue indices of $H$, let $V_S$ hold corresponding orthonormal eigenvectors, and let $W\in\R^{n\times|S|}$ have orthonormal columns. For $\Phi(G)=\nuc{G}\msign(G)$ and $E_0=cV_SW^\top$ with $c\ne0$, the iteration \eqref{eq:iteration} gives $E_t=(1-\alpha\sum_{i\in S}\lambda_i)^tE_0$ for every $t\ge0$.
\end{proposition}

\begin{proof}
For $E=cV_SW^\top$ with $c>0$, $HE=cV_S\Lambda_SW^\top$ with $\Lambda_S=\diag(\lambda_i)_{i\in S}$, which is a thin SVD since $V_S$ and $W$ have orthonormal columns and $c\lambda_i>0$. Hence $\msign(HE)=V_SW^\top$, $\nuc{HE}=c\sum_{i\in S}\lambda_i$, and $E-\alpha\Phi(HE)=(1-\alpha\sum_{i\in S}\lambda_i)E$. For $c<0$ apply the same to $-E=|c|V_S(-W)^\top$. By induction, $E_t=(1-\alpha\sum_{i\in S}\lambda_i)^tE_0$.
\end{proof}

\emph{Exactness of the threshold.} Along such a direction the iteration is a scalar GD with curvature $\sum_{i\in S}\lambda_i$. Taking $S$ to be the indices of the $n$ largest eigenvalues, so that $\sum_{i\in S}\lambda_i=\tr_n(H)$, gives a two-cycle at $\alpha\tr_n(H)=2$ and divergence above it, so the threshold of Theorem~\ref{thm:nuclear} with $\tr_n(H)$ cannot be relaxed.

\emph{Dependence of the rate on the initialization.} Taking $S=\{i\}$ gives the rate $|1-\alpha\lambda_i|$, which is that of GD, as it must be, since $\nuc{G}\msign(G)=G$ for a rank-one gradient. The asymptotic rate therefore depends on the initialization, and from a direction with $\alpha\sum_{i\in S}\lambda_i=1$ the minimizer is reached in one step. For a general initialization the iteration is a nonlinear map on the directions of the error, since the update is homogeneous of degree one in the gradient, and we do not know its rate in closed form. In experiments on random problems (not reported here), the measured asymptotic factor is typically smaller than the slowest invariant rate $\max_S|1-\alpha\sum_{i\in S}\lambda_i|$, and it varies with the initialization.

\begin{remark}
The identity \eqref{eq:nuclear-identity} is exact, so at a fixed iterate the loss decreases if and only if $\alpha\tr(U_t^\top HU_t)<2$, and the largest one-step decrease, at $\alpha=1/\tr(U_t^\top HU_t)$, equals $\nuc{G_t}^2/(2\tr(U_t^\top HU_t))$. The quantity $\tr(U_t^\top HU_t)$ is the trace of $H$ on the span of the current gradient, and $\tr_n(H)$ is its worst case over all spans of dimension at most $n$. Equivalently, $\tr_n(H)=\sup\{\ip{D}{HD}:\opnorm{D}\le1\}$, which equals $\tr(H)$ for $n\ge d$, is the smoothness constant of $f$ in the geometry of the spectral norm, in which the polar factor is the steepest descent direction.
\end{remark}

\section{Proof of the nonconvergence lemma}
\label{app:nonconvergence}

We identify $\R^{d\times n}$ with $\R^N$, $N=dn$, and call a set null if it has Lebesgue measure zero. A cone is a set $\mathcal Z$ with $t\mathcal Z=\mathcal Z$ for all $t>0$.

The proof has two parts. First, an update that depends only on the direction of the gradient moves the iterate by at least $\alpha\gamma$ at every step, so a trajectory can converge only by reaching $X_\star$ exactly after finitely many steps. Second, the set of initializations that do so is null. This is not automatic, since the step map $E\mapsto E-\alpha\Phi(HE)$ is not injective and, for the polar map, not differentiable at rank-deficient gradients. Lemma~\ref{lem:null} reduces the claim to showing that the critical set of the step map is null, and Step~3 of the proof of Proposition~\ref{prop:nonconv-general} shows that this set meets every ray in finitely many points.

\begin{lemma}[Null preimages]
\label{lem:null}
Let $S:\R^N\to\R^N$ be continuously differentiable on the complement of a closed null set $\mathcal Z$, and let $\mathcal C=\{E\notin\mathcal Z:\det DS(E)=0\}$. If $\mathcal C$ is null, then $S^{-1}(\mathcal A)$ is null for every null set $\mathcal A$.
\end{lemma}

\begin{proof}
Every $E\notin\mathcal Z\cup\mathcal C$ has, by the inverse function theorem, an open neighborhood on which $S$ is a $C^1$ diffeomorphism onto its image, and on a smaller neighborhood with compact closure the inverse is Lipschitz. A Lipschitz map sends null sets to null sets, so the part of $S^{-1}(\mathcal A)$ in that neighborhood is null. The open set $\R^N\setminus(\mathcal Z\cup\mathcal C)$ is covered by countably many such neighborhoods, so $S^{-1}(\mathcal A)\setminus(\mathcal Z\cup\mathcal C)$ is a countable union of null sets. The rest of $S^{-1}(\mathcal A)$ lies in the null set $\mathcal Z\cup\mathcal C$.
\end{proof}

\begin{proposition}[Nonconvergence of direction-only updates]
\label{prop:nonconv-general}
Let $\Phi:\R^{d\times n}\to\R^{d\times n}$, $\Phi(0)=0$, satisfy: (H1) $\Phi(tG)=\Phi(G)$ for all $G\ne0$ and $t>0$ (direction only); (H2) $\gamma:=\inf\{\Fro{\Phi(G)}:\Fro{G}=1\}>0$ (length bounded below); (H3) there is a closed null cone $\mathcal Z$ such that $\Phi$ is continuously differentiable on $\R^{d\times n}\setminus\mathcal Z$ (smooth off a null cone). Then for every $H\succ0$ and $\alpha>0$, the sequence $E_{t+1}=E_t-\alpha\Phi(HE_t)$ satisfies $\max\{\Fro{E_t},\Fro{E_{t+1}}\}\ge\alpha\gamma/2$ whenever $E_t\ne0$, and it has no limit for almost every $E_0$.
\end{proposition}

\begin{proof}
\emph{Step 1.} \emph{The increments are bounded below.} If $G_t=HE_t\ne0$, then by (H1) and (H2), $\Fro{E_{t+1}-E_t}=\alpha\Fro{\Phi(G_t/\Fro{G_t})}\ge\alpha\gamma$. The triangle inequality gives $\Fro{E_t}+\Fro{E_{t+1}}\ge\alpha\gamma$, hence $\max\{\Fro{E_t},\Fro{E_{t+1}}\}\ge\alpha\gamma/2$. Moreover, a convergent sequence has vanishing increments, so if $E_t$ converges then $E_t=0$ for some $t$. Hence the convergent initializations lie in $\bigcup_{K\ge0}S^{-K}(\{0\})$, where $S(E)=E-\alpha\Phi(HE)$.

\emph{Step 2.} \emph{Smoothness of the step map.} Put $\mathcal Z'=\{E:HE\in\mathcal Z\}$. Since $E\mapsto HE$ is an invertible linear map, $\mathcal Z'$ is a closed null cone, and $S$ is $C^1$ on its complement.

\emph{Step 3.} \emph{The critical set is null.} For $E\notin\mathcal Z'$ let $J(E)=D_E[\Phi(HE)]=D\Phi(HE)H$, so $DS(E)=I-\alpha J(E)$. Differentiating (H1) at $tG$ gives $D\Phi(tG)=t^{-1}D\Phi(G)$, hence $J(tE)=t^{-1}J(E)$ for $t>0$. Along the ray through $E$, $\det DS(tE)=\det(I-\alpha t^{-1}J(E))$ is a polynomial in $1/t$ of degree at most $N$ with constant term $1$, so it has finitely many positive roots. Thus $\mathcal C=\{E\notin\mathcal Z':\det DS(E)=0\}$, which is relatively closed and hence measurable, meets every ray in finitely many points, and by Fubini's theorem in polar coordinates $\mathcal C$ is null.

\emph{Step 4.} \emph{Conclusion.} By Lemma~\ref{lem:null}, preimages of null sets under $S$ are null. Since $\{0\}$ is null, every $S^{-K}(\{0\})$ is null by induction, and so is their countable union.
\end{proof}

\begin{proof}[Proof of Lemma~\ref{thm:nonconv}]
We verify (H1) to (H3) for the two instances and identify $\gamma$. Proposition~\ref{prop:nonconv-general} then gives both statements of the lemma, since $E_t\ne0$ if and only if $G_t\ne0$.

\emph{Polar map.} (H1) holds since the polar factor of $tG$ is that of $G$. (H2) holds with $\gamma=1$ since $\Fro{UV^\top}^2=\tr(VU^\top UV^\top)=r\ge1$. For (H3), let $\mathcal Z$ be the set of matrices of rank below $\min(d,n)$. It is a closed cone, and it is null as the zero set of the nonzero polynomial $\det(G^\top G)$ (if $d\ge n$) or $\det(GG^\top)$ (if $d<n$). On its complement, $\msign(G)=G(G^\top G)^{-1/2}$ or $(GG^\top)^{-1/2}G$, which is $C^\infty$ since $A\mapsto A^{-1/2}$ is $C^\infty$ on positive definite matrices.

\emph{Normalized NS iteration with $\varepsilon=0$.} Write $p(u)=\sum_ja_{2j+1}u^{2j+1}$. For $Y=U\Sigma V^\top$ one has $Y(Y^\top Y)^j=U\Sigma^{2j+1}V^\top$, hence
\[
\Phi(G)=U\diag\big(p(\sigma_i/\Fro{G})\big)V^\top=\sum_ja_{2j+1}\frac{G(G^\top G)^j}{\Fro{G}^{2j+1}},\qquad G\ne0,
\]
which is $C^\infty$ on $\R^{d\times n}\setminus\{0\}$, including at rank-deficient matrices, and satisfies (H1). So (H3) holds with $\mathcal Z=\{0\}$. For (H2), on the unit sphere the singular values $u_i=\sigma_i/\Fro{G}$ satisfy $\sum_iu_i^2=1$, and $p(u)\ge c_pu$ on $(0,1]$ gives $\Fro{\Phi(G)}^2=\sum_ip(u_i)^2\ge c_p^2\sum_iu_i^2=c_p^2$, so $\gamma\ge c_p>0$, and the floor holds with $\gamma=c_p$. The argument uses only that $p$ is an odd polynomial with $p'(0)>0$ that is positive on $(0,1]$, so it holds for the NS polynomial of every degree.
\end{proof}

\section{Boundary behavior and a vanishing length that does not suffice}
\label{app:boundary}

This appendix uses a diagonal model, in which the iteration decouples into scalar recursions, to characterize exactly when a fixed shaping converges (Proposition~\ref{prop:diagonal}), to quantify the slow convergence at the boundary $\alpha C\lambda_{\max}=2$ of Theorem~\ref{thm:convergence}, and to show that a vanishing length alone does not give convergence (Proposition~\ref{prop:power}). Take $d=n$, $H=\diag(\lambda_i)$, and a diagonal error $E_t=\diag(x_{t,i})$. Then $G_t$ is diagonal with singular values $\lambda_i|x_{t,i}|$, and a fixed shaping $g_i=h(\sigma_i)$ acts on every signed coordinate as
\begin{equation}
x_{t+1,i}=x_{t,i}\big(1-\alpha\lambda_i\,r_h(\lambda_i|x_{t,i}|)\big),\qquad r_h(\sigma)=h(\sigma)/\sigma.
\label{eq:scalar}
\end{equation}
The same recursion describes a rank-one error $E_t=x_twv^\top$ with $Hw=\lambda w$. A general error matrix does not decouple in this way, which is why the following statements are restricted to the diagonal model.

\begin{proposition}[Diagonal characterization]
\label{prop:diagonal}
Let $h$ be bounded, continuous and positive on $(0,\infty)$ with $\lim_{\sigma\downarrow0}h(\sigma)/\sigma=a\in(0,\infty)$, and let $C_h=\sup_{\sigma>0}h(\sigma)/\sigma$. In the diagonal model, every initialization converges to zero if and only if either $\alpha C_h\lambda_{\max}<2$, or $\alpha C_h\lambda_{\max}=2$ and $r_h(\sigma)<C_h$ for every $\sigma>0$.
\end{proposition}

\begin{proof}
The strict inequality gives convergence by Theorem~\ref{thm:convergence}, since the gains on any range are at most $C_h$.

Suppose $\alpha C_h\lambda_{\max}>2$. The gain $r_h$ is continuous on $(0,\infty)$, tends to $a$ at zero and to $0$ at infinity because $h$ is bounded, and exceeds $2/(\alpha\lambda_{\max})<C_h$ somewhere. By the intermediate value theorem there is $\sigma_2>0$ with $r_h(\sigma_2)=2/(\alpha\lambda_{\max})$. The initialization with top coordinate $x=\sigma_2/\lambda_{\max}$ and all other coordinates zero has the multiplier $-1$ in \eqref{eq:scalar}, so it alternates between $x$ and $-x$ forever.

At equality, if the supremum $C_h$ is attained at some $\sigma_2$, the same two-cycle exists. If it is not attained, then $0<\alpha\lambda_ir_h(\sigma)<2$ for every $\sigma>0$ and every $i$, so every nonzero coordinate strictly decreases in magnitude. Let $|x_{t,i}|\downarrow L$. If $L>0$, the multiplier tends to $1-\alpha\lambda_ir_h(\lambda_iL)$, which has modulus strictly below one, contradicting $|x_{t+1,i}|/|x_{t,i}|\to1$. Hence $L=0$.
\end{proof}

\emph{Rate at the boundary.} For a bounded $h$, an unattained supremum must equal $a$, since $r_h\to0$ at infinity and a continuous function attains its maximum on a compact interval. At equality the multiplier of the top coordinate then tends to $-1$, and the convergence is sublinear: if $h(\sigma)=a\sigma-b\sigma^{1+\nu}+o(\sigma^{1+\nu})$ with $b,\nu>0$ and $\alpha a\lambda_{\max}=2$, then a small top-coordinate magnitude $y_t=|x_t|$ obeys $y_{t+1}=y_t-\alpha b\lambda_{\max}^{1+\nu}y_t^{1+\nu}+o(y_t^{1+\nu})$, so that $y_t=\Theta(t^{-1/\nu})$. Both $\tanh(\sigma/\theta)$ and $\sigma/\sqrt{\sigma^2+\epsilon^2}$ have $\nu=2$, hence an error of order $t^{-1/2}$ and a loss of order $t^{-1}$ at the boundary. This is the slow decay seen at equality in experiment P1.

\begin{proposition}[Fractional powers cycle]
\label{prop:power}
Let $d=n$, $H=\diag(\lambda_i)$, $g_i=\sigma_i^\gamma$ with $0<\gamma<1$, and let $E_0$ be diagonal. Then $E_t$ stays diagonal, the length $\Fro{\Phi(G)}$ tends to zero as $E\to0$, and yet, for every $\alpha>0$, $E_t\to0$ holds only for $E_0$ in a null set of diagonal matrices.
\end{proposition}

\begin{proof}
For $E=\diag(x_i)$, $G=\diag(\lambda_ix_i)$ has singular values $\lambda_i|x_i|$ with singular vectors $\mathrm{sign}(x_i)e_i$ and $e_i$, so $\Phi(G)=\diag\big((\lambda_i|x_i|)^\gamma\mathrm{sign}(x_i)\big)$ and the coordinates evolve independently as $x\mapsto T_i(x)=x-c_i|x|^\gamma\mathrm{sign}(x)$ with $c_i=\alpha\lambda_i^\gamma>0$. The length $(\sum_i(\lambda_i|x_i|)^{2\gamma})^{1/2}$ tends to zero as $E\to0$. Fix $i$, drop the index, and let $y_\star=(c/2)^{1/(1-\gamma)}$. For $0<|x|<y_\star$ we have $c|x|^\gamma>2|x|$, so $T(x)$ has the opposite sign to $x$ and $|T(x)|=c|x|^\gamma-|x|>|x|$. If $x_t\to0$ with $x_t\ne0$ for all $t$, then $|x_t|<y_\star$ for $t\ge t_0$, hence $|x_{t+1}|>|x_t|$ for $t\ge t_0$ and $|x_t|\ge|x_{t_0}|>0$, a contradiction. So $x_t\to0$ forces $x_t=0$ for some $t$. The map $T$ is real-analytic and nonconstant on $(0,\infty)$ and on $(-\infty,0)$, so every level set $T^{-1}(\{y\})$ is countable, and by induction so is $\mathcal N=\bigcup_{k\ge0}T^{-k}(\{0\})$, the set of $x_0$ from which $0$ is reached in finitely many steps. Since $E_t\to0$ requires every coordinate to converge, the set of diagonal $E_0$ with $E_t\to0$ is contained in $\{E_0:x_{0,1}\in\mathcal N_1\}$, which is null.
\end{proof}

Proposition~\ref{prop:power} shows that a vanishing length is necessary but not sufficient for convergence at a constant step size. In the diagonal model, the update must vanish at least linearly in the gradient, which is what a bounded gain provides directly, and what the nuclear scaling provides through the factor $\nuc{G}$. In the scalar model, the degree of homogeneity of the update decides the outcome: degree zero (polar map, normalized NS) gives no convergence, degree in $(0,1)$ gives cycles of size $(\alpha\lambda^\gamma/2)^{1/(1-\gamma)}$, degree one allows linear convergence, and degree above one converges only sublinearly.

\section{The Newton--Schulz stages of Muon}
\label{app:muon}

The stage polynomial of the reference implementation~\cite{jordan2024muon} is $q(u)=3.4445u-4.7750u^3+2.0315u^5$, applied $k=5$ times, and the composed polynomial is $p=q^{\circ5}$. The stage gain $q(u)/u=3.4445-4.7750u^2+2.0315u^4$ is a quadratic in $u^2$ with a negative discriminant, so $q(u)>0$ for every $u>0$, and hence $p>0$ on $(0,\infty)$. On $0\le u\le1.25$ the stage gain is at most $3.4445$, and the endpoint and stationary values of $q$ show that $q$ maps $[0,1.25]$ into $[0,1.2024]$. Hence every stage that starts in $[0,1]$ stays in $[0,1.25]$, and $q^{\circ5}(u)/u\le3.4445^5$ for $0<u\le1$, with equality in the limit $u\downarrow0$ since $(q^{\circ5})'(0)=3.4445^5$ by the chain rule. Therefore $C_p=p'(0)=3.4445^5\approx484.9$, and a numerical evaluation gives $c_p\approx0.696$ and $\max_{0\le u\le1}p(u)\approx1.202$. The same computation for $k=1,2,3$ gives $C_p=3.4445^k$ in every case.

With these constants, the guarantee of Section~\ref{sec:bounded} for the normalized NS iteration with the default offset $\varepsilon=10^{-7}$ requires $\alpha\lambda_{\max}<2\varepsilon/3.4445^5\approx4\times10^{-10}$. The offset therefore places the update in the bounded-gain regime only formally. For every step size used in practice, and for as long as $\Fro{G_t}\gg\varepsilon$, the normalized singular values $\sigma_i/(\Fro{G_t}+\varepsilon)$ are indistinguishable from $\sigma_i/\Fro{G_t}$, and the iteration behaves as the case $\varepsilon=0$ of Lemma~\ref{thm:nonconv}. The same holds for the offset $10^{-6}$ used in the experiments of~\cite{gonon2026insights}, whose constant-step Newton--Schulz trajectories plateau like the exact ones (see their Fig.~10, where the input is divided by $1.02\Fro{G}+10^{-6}$). Adding stages does not change this picture: it raises $C_p$ by a factor $3.4445$ per stage and shrinks the guaranteed range accordingly. Note that $C_p$ measures the amplification of the polynomial and not its distance to the polar factor.

\section{Experimental details}
\label{app:experiments}

The target is $X_\star=0$, so every loss is evaluated directly from the error. All experiments use NumPy $2.4.2$ and Matplotlib $3.10.8$.

\noindent\textbf{Diagonal experiments (P1, P2).}
The Hessian eigenvalues are $\lambda_i=30^{-1}\,30^{(i-1)/15}$ for $i=1,\ldots,16$, and $E_0=I/4$. The threshold experiment uses 1200 iterations at $\alpha\lambda_{\max}\in\{1.8,2.0,2.2\}$ for $h(\sigma)=\tanh\sigma$, whose gain is at most $C=1$. The rate experiment uses 15 equally spaced effective step sizes $\alpha a$ from $0.3$ to $1.98$, where $a=h'(0)$, for GD, $\tanh\sigma$, $\tanh4\sigma$, and $\sigma/\sqrt{\sigma^2+0.25}$. A least-squares fit of $\log\Fro{E_t}$ against $t$ over the final 300 of 2400 updates gives the measured rate. The deviation from $\rho(\alpha)$ is $2.5\times10^{-8}$ at $\alpha a=0.3$, $4.5\times10^{-14}$ at $0.66$, and within one unit in the last place for $\alpha a\ge0.9$. The sampled grid brackets the optimal effective step size $60/31\approx1.94$ without containing it, so the smallest sampled factor is $0.938$ at $\alpha a=1.86$.

\noindent\textbf{Dense experiment (P3).}
A NumPy default generator with seed $20260912$ draws a $16\times16$ Gaussian matrix whose QR factor $Q$ defines $H=Q\diag(\lambda_i)Q^\top$ with the eigenvalues above, so that $\tr(H)\approx4.798$ and $\tr_8(H)\approx4.126$, and then a $16\times8$ Gaussian error that is divided by its Frobenius norm. The recorded $H$ and $E_0$ allow an exact reconstruction. Four runs share the step size $\alpha=0.01143446$ and the 1400-step horizon: the normalized NS iteration with $\varepsilon=0$, the normalized NS iteration with $\varepsilon=3.080166$ (for which $\alpha C_p\lambda_{\max}/\varepsilon=1.8$ lies inside the range of Theorem~\ref{thm:convergence}), the exact polar map by a thin SVD, and GD. Both NS runs apply five copies of the stage $Y\leftarrow Y(3.4445I-4.7750Y^\top Y+2.0315(Y^\top Y)^2)$, and the matrix-product and SVD evaluations of the polynomial agree to $5\times10^{-15}$. PolarGrad runs from the same $E_0$ at $\alpha=u/\tr(H)$ for $u\in\{1.0,1.9,2.4\}$, which corresponds to $\alpha \operatorname{tr}_8(H) \in \{0.86, 1.63, 2.06\}$. The runs for $u = 1.0$ and $1.9$ lie inside the exact stability range $\alpha \operatorname{tr}_8(H) < 2$ of Appendix~\ref{app:nuclear}, while $u = 2.4$ exceeds it.

\emph{Recorded values.} Because the polar and the $\varepsilon=0$ trajectories oscillate, the recorded summary is the range of the relative loss over the final 300 steps: $1.02\times10^{-4}$ to $1.16\times10^{-3}$ for the $\varepsilon=0$ run and $2.94\times10^{-4}$ to $5.97\times10^{-4}$ for the polar run, against $1.14\times10^{-2}$ to $1.46\times10^{-2}$ for GD. The $\varepsilon=3.08$ run reaches $9.65 \times 10^{-76}$ and the PolarGrad runs reach $2.30 \times 10^{-17}$ and $3.01 \times 10^{-82}$ at $u = 1.0$ and $1.9$ by the final step, all still decreasing. The PolarGrad run at $u = 2.4$, however, fails to converge and oscillates with a minimum relative loss of $5.48 \times 10^{-1}$, consistent with Proposition~\ref{prop:invariant}, which shows that the threshold $\alpha\tr_8(H)<2$ cannot be relaxed.

\noindent\textbf{Offset experiment (P4).}
The normalized NS iteration runs on the same $H$ and the same unit direction of $E_0$ at nine pairs formed from $\varepsilon\in\{10^{-1},3\times10^{-2},10^{-2}\}$ and $\alpha p'(0)\lambda_{\max}/\varepsilon\in\{1.0,1.9,2.05\}$, with $p'(0)=3.4445^5$. Every run uses $\Fro{E_0}=10^{-12}\varepsilon$, so that the iteration is in the local regime of Proposition~\ref{prop:rate} throughout, and the largest gradient norm reached is below $6.7\times10^{-5}\varepsilon$. Every trajectory runs for 400 steps, and the factor is the exponential of a least-squares slope of $\tfrac12\log f(X_t)$ over the final 200. The spread of the measured factor at matched $\alpha p'(0)/\varepsilon$ is $2.2\times10^{-16}$, and the deviation from the predicted factor is $2.31\times10^{-5}$, with measured values $0.967$, $0.937$, and $1.050$.